\documentclass[journal]{IEEEtran}
\ifCLASSINFOpdf

\else

\fi

\PassOptionsToPackage{numbers}{natbib}

\usepackage{multirow}

\usepackage{amsmath}

\usepackage{graphicx}

\usepackage{color}

\newcommand{\tabincell}[2]{\begin{tabular}{@{}#1@{}}#2\end{tabular}}

\begin{document}

\title{Spatial-Temporal Recurrent Neural Network for Emotion Recognition}

\author{Tong~Zhang,
        Wenming~Zheng*,~\IEEEmembership{Member,~IEEE,}
        Zhen~Cui*,
        Yuan~Zong
        and Yang Li
\thanks{Tong Zhang and Yang Li are with the Key Laboratory of Child Development and Learning Science of Ministry of Education, and the Department
of Information Science and  Engineering, Southeast University, China. (
e-mail: tongzhang@seu.edu.cn;yang\_li@seu.edu.cn).}
\thanks{Wenming Zheng, Zhen Cui and Yuan Zong are with the Key Laboratory of Child Development and Learning Science of Ministry of Education, Research Center for Learning Science, Southeast University, Nanjing, Jiangsu 210096, China (e-mail: wenming\_zheng@seu.edu.cn; zhen.cui@seu.edu.cn; xhzongyuan@seu.edu.cn). \protect\\ 
Asterisk indicates corresponding author.\protect\\
}
}

\markboth{Journal of \LaTeX\ Class Files,~Vol.~13, No.~9, September~2014}%
{Shell \MakeLowercase{\textit{et al.}}: Bare Demo of IEEEtran.cls for Journals}

\maketitle

\begin{abstract}
Emotion analysis is a crucial problem to endow artifact machines with real intelligence in many large potential applications. As external appearances of human emotions, electroencephalogram (EEG) signals and video face signals are widely used to track and analyze human's affective information. According to their common characteristics of spatial-temporal volumes, in this paper we propose a novel deep learning framework named spatial-temporal recurrent neural network (STRNN) to unify the learning of two different signal sources into a spatial-temporal dependency model. In STRNN, to capture those spatially co-occurrent variations of human emotions, a multi-directional recurrent neural network (RNN) layer is employed to capture long-range contextual cues by traversing the spatial region of each time slice from multiple angles. Then a bi-directional temporal RNN layer is further used to learn discriminative temporal dependencies from the sequences concatenating spatial features of each time slices produced from the spatial RNN layer. To further select those salient regions of emotion representation, we impose sparse projection onto those hidden states of spatial and temporal domains, which actually also increases the model discriminant ability because of this global consideration. Consequently, such a two-layer RNN model builds spatial dependencies as well as temporal dependencies of the input signals. Experimental results on the public emotion datasets of EEG and facial expression demonstrate the proposed STRNN method is more competitive over those state-of-the-art methods.
\end{abstract}


\IEEEpeerreviewmaketitle

\section{Introduction}

\IEEEPARstart{H}{uman} emotion analysis plays a crucial role in endowing artifact machines with humanized characteristics, and is arousing more and more attentions due to its potential applications to human-machine interaction. The recent advances of  AlphaGo~\cite{silver2016mastering} still lacks of the ability of understanding of human emotions or human-like emotion interaction, though it is able to beat the apex master. Of course, the study of emotion analysis has faced significant difficulties due to its intrinsic property of less tangibility. Nevertheless, it is notable that emotions can be conveyed in external signal forms by utilizing some electric devices. Especially, with the development of hardware techniques, it is becoming easy to collect signals reflecting human emotions, such as acoustic waves, facial video sequences, electroencephalogram (EEG) signals and so on.

Among the various emotion signals, EEG and facial expression images are widely employed for emotion analysis. For EEG, signals are produced from multiple active electrodes attached on cerebral cortex by arranging a certain spatial layout. To deal with EEG signals, Zheng et al.~\cite{zheng2015investigating}  concatenated the multi-channel signals and fed them into  deep belief networks (DBNs) to perform emotion recognition. Likewise, facial expression signals can be collected by using general cameras. For facial expression recognition, many methods are proposed to learn more robust features by using the state-of-the-art deep learning techniques such as convolutional neural network (CNN) \cite{krizhevsky2012imagenet, parkhi2015deep} and recurrent neural network (RNN) \cite{williams1989learning}. However, most existing emotion recognition methods based on EEG signals or facial expression images ignore spatial dependencies, i.e., the relationship between multiple electrodes or the relationships between facial local areas. Building these relationships can actually produce some bundling characteristics reflecting human emotions, while the behaviors of emotions are not usually isolate. Furthermore, these emotion related signals contain not only spatial components at a single moment but also contextual dependencies among temporal slices. In order to better recognize human emotion, the crucial spatial and temporal dependencies should be well modeled. However, most of the previous methods consider only either spatial feature learning or temporal dependency construction.

In this paper, we propose a unified deep network framework called spatial-temporal recurrent neural network (STRNN) to deal with both EEG based emotion recognition and facial emotion recognition. STRNN can not only learn spatial dependencies of multi-electrode or image context itself, but also learn a long-term memory information in temporal sequences. To learn spatial dependencies, a quad-directional spatial RNN (SRNN) layer is employed to scan each temporal slice by adopting a certain spatial orders respectively from different angles, and finally produce a discriminative dependency sequence in the slice. Then a bi-directional temporal RNN layer (TRNN) is further stacked on SRNN to capture long-term temporal dependencies by scanning the temporal sequences forward and backward. In each RNN layer, the previous states are connected to the current one so that the sub-network is inherently deep and able to retain all the past inputs and discover correlations among the data that might be far away from each other. The benefit of the hierarchical RNN is that the two layers may act as two memory units to remember and encode all the scanned spatial and temporal area so that the proposed STRNN is able to globally model the spatial and temporal dependencies.

As emotion stimuli are usually activated in some local regions, we expect to discover those salient stimulus regions expressing human emotions. To this end, we introduce the sparse projection transformation onto those spatially encoding states to detect those salient activation points. Besides, as the global projection is operated on the entire spatial domain, the learnt STRNN can automatically bundle those co-occurrent emotion regions, which would bring some gains for the final emotion recognition. Similarly, the sparse projection is also used for those temporally encoded states in order to adaptively choose and combine those time slices with more discriminability.

The main contributions of our work can be summarized as follows:
\begin{enumerate}
\item[(1)] We propose an end-to-end STRNN to jointly integrate spatial and temporal dependencies, as well as learn discriminative features. To the best our knowledge, this would be the first  model that utilizes both spatial and temporal information in RNN.
\item[(2)] Two emotion recognition tasks, i.e., EEG based emotion recognition and facial emotion recognition, are investigated and unified under a deep network framework by constructing spatial-temporal volumes, where EEG signals are spatially organized in electrode coordinations.
\item[(3)] Salient emotion activation regions can be effectively captured by introducing sparse projection on those encoding hidden states, which can naturally bundle those co-occurred emotion activation regions by adaptively weighting them.
\end{enumerate}

The paper is organized as follows: Section \ref{relawork} overview some approaches related to RNN  and  human emotion recognition in EEG signals and videos; in Section \ref{modeldes} we proposed the STRNN method for emotion recognition in details; We present  experimental results to evaluate the proposed method in Section \ref{Exp} and finally in Section \ref{conclu} we conclude the paper.

\begin{figure*}[!t]
  \centering
  \includegraphics[height=2.8in,width=7in]{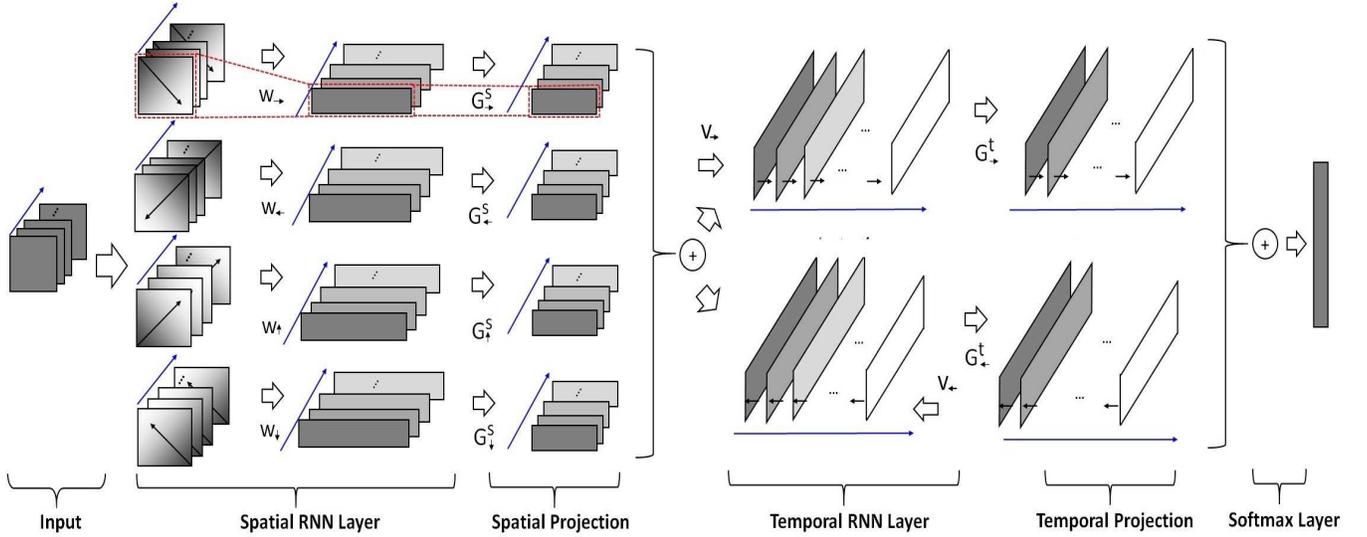}
  \caption{An overview on the proposed STRNN framework. The SRNN and TRNN are elaborately integrated and jointly learnt to capture spatial and temporal dependencies. The blue arrow indicates the temporal axis. More details can be found in Section \ref{modeldes}.}
\label{frameworkpic}
\end{figure*}
\section{Related work}
\label{relawork}
Human emotion recognition based on EEG signals or facial expression images had been extensively investigated during the past decades, and a lot of algorithms have been proposed in the literatures to this end. For instances, for EEG based emotion recognition, descriptors such as high order crossings \cite{petrantonakis2010emotion} and differential entropy (DE) \cite{duan2013differential} are employed, and  popular classifiers such as support vector machine (SVM) \cite{burges1998tutorial} and group sparse canonical correlation analysis (GSCCA) \cite{zhengmultichannel} are used to achieve classification. In \cite{zheng2015investigating}, DBNs are used to  learn high-level features from the extracted DE descriptors.  On the other hand, for facial expression images based emotion recognition, various hand-crafted facial features has been applied such as 3D HOG \cite{klaser2008spatio}, 3D SIFT \cite{scovanner20073}, expressionlet-based spatio-temporal manifold (STM-ExpLet) \cite{liu2014learning} and so on. In addition, deep learning methods are also employed to deal with the expression recognition problem from facial image sequences in recent years, e.g., the 3DCNN method proposed in \cite{liu2014deeply}. The method named 3DCNN-DAP proposed  in \cite{liu2014deeply} employs 3DCNN  while using the strong spatial structural constraints on the dynamic action parts to extract robust representation from videos.

Recently, RNNs have achieved great success in processing sequential data such as natural language processing \cite{graves2014towards, mikolov2010recurrent}, action recognition \cite{du2015hierarchical}, speech recognition \cite{graves2013speech} and so on. Then RNN is further improved  to deal with images \cite{zuo2015convolutional} by scanning the regions of images into sequences in certain directions. Due to the ability of retaining information about the past inputs, RNN is able to learn contextual dependencies with images, which is advantageous compared to CNN. This is because CNN may fail to learn the global dependencies due to the locality of convolution and pooling layers. For this reason, RNN is usually  combined with CNN in order to achieve better performance of vision tasks such as image segmentation \cite{visin2015reseg} and recognition \cite{visin2015renet}.

In what follows, we will propose our STRNN method, which uses an end-to-end spatial-temporal learning network of to simultaneously integrate spatial and temporal dependencies for co-adaptively dealing with EEG based emotion recognition and video based facial expression recognition.

\section{STRNN Model for Emotion Recognition}
\label{modeldes}

To specify the proposed STRNN method, we illustrate the framework of STRNN in Fig.~\ref{frameworkpic}, in which the inputs of networks would be any spatial-temporal style data such as multi-channel sequence signals (e.g., EEG) or spatial-temporal cubic volumes (e.g., videos) as long as  they could be traversed in a certain orders in space or time domain.
In Fig.~\ref{frameworkpic}, we take cubic videos as an example for the simplicity of description. The goal of the proposed framework is to capture spatial-temporal information within sequence signals. To realize this point, we attempt to stack two-layer RNNs, i.e., an SRNN layer and a TRNN layer, so as to concatenating other layers for an end-to-end deep neural network.
Consequently, STRNN combines spatial and temporal information simultaneously by building the dependencies of adjacent and even long-term elements. Moreover, to detect those salient emotion regions, sparse projections are further applied to those encoding hidden states of SRNN and TRNN layers.

To model spatial dependencies on each time slice, i.e., the relationships between responses of multiple electrodes at a certain moment for EEG signals, we use RNN to scan all spatial elements under a predefined order strategy. Unlike those Markov chain structures frequently used for such a graph model, RNN simplifies its process by unfolding graph structures into an order structure, which makes the learning more controllable. After scanning the time slice element by element sequentially, RNN can characterize those low-level elements and their complex dependencies if long-term recurrent units (e.g., long short-term memory, LSTM~\cite{hochreiter1997long}) are adopted. However, it is notable that the data could be contaminated by signal noises in EEG or partial occlusion in videos, a single RNN used in a 2D space  may not be enough to resist these variations. For this problem, we use four directional RNNs to traverse the spatial region at a time slice from four specific angles. The four RNNs are actually complementary for constructing a complete relationship graph, and thus alleviate the effect of noises while simplifying those techniques used for modeling graph structures. Concretely, when modeling spatial dependencies, we use a graph $\mathcal{G}_t=\{\mathcal{R}_t,\mathcal{C}_t\}$ to represent the spatial elements in the $t$th slice denoted by $\mathbf{X}_t$, where $\mathcal{R}_t=\{\mathbf{x}_{tij}\}(i=1,\cdots,h, j=1,\cdots,w)$ represents the vertex set of spatial elements indexed by their spatial coordinates, and $\mathcal{C}_t=\{e_{tij,tkl}\}$ denotes the edges of spatial neighboring elements in the $t$th slice. Then we traverse through $\mathcal{G}_t$  with a predefined forward evolution sequence so that the input state and previous states can be defined for a RNN unit.  Formally, the adopted multidirectional spatial RNNs in STRNN can be written as
\begin{equation} \label{eq1}
  \mathbf{h}_{tij}^r=\sigma_1(\mathbf{U}^r\mathbf{x}_{tij}+\sum_{k=1}^h\sum_{l=1}^w\mathbf{W}^r\mathbf{h}^r_{tkl}\times e_{tij,tkl} +\mathbf{b}^r)  \\
\end{equation}

\begin{equation} \label{eq1_add}
            e_{tij,tkl}=
                  \begin{cases}
                  1,  &  \mbox{if}~~(k,l) \in \mathcal{N}_{ij}; \\
                  0,  &  \mbox{otherwise.}
                  \end{cases}
\end{equation}
where $\mathbf{x}_{tij}$ and  $\mathbf{h}_{tij}^r$ respectively denote the representation of input and  hidden  node at the location of $(i, j)$  in the $t$th slice, $\mathcal{N}_{ij}$ is the set of predecessors of the vertex $(i, j)$ and $r$ represents a certain traversing direction. Hence, $\mathbf{h}_{tij}^r$ collects information of all the previous scanned elements of the current state $(i, j)$. The non-linear function denoted by $\sigma_1(\cdot)$ for hidden layers is ReLU or Sigmoid function. As this SRNN layer traverses all the vertexes in $\mathcal{R}_t$,  the number of the hidden states  equals $h \times w$ for a given traversing direction. For simplification, the output hidden states  denoted as $ \mathbf{h}_{tij}^r (i=1,\cdots,h, j=1,\cdots,w)$ are rewritten as $\mathbf{h}_{tk}^r (k=1,\cdots,K)$, where $K$  equals $h \times w$. To further detect those salient regions of emotion representation,  projection  matrices  are applied to  the spatial  hidden states corresponding to different traversing directions. Assume that a projection  matrix for a certain  traversing direction is denoted as  $\mathbf{G}^{r}=[G^r_{ij}]_{K \times K_p}$ where $K_p$ denotes the number of the hidden states after projection, then the projection can be written as
\begin{equation} \label{sppro}
  \mathbf{s}^r_{tl}=\sum_{i=1}^KG^r_{il}\mathbf{h}_{ti}^r, l=1,\cdots,K_p
\end{equation}
where $\mathbf{s}^r_{tl}$ denotes the $l$th column vector of the output matrix of the projection. Let $ \mathbf{s}^r_{t}=[(\mathbf{s}^r_{t1})^T, \cdots,  (\mathbf{s}^r_{tK_p})^T ]^T$  denote the concatenated vector of the column vectors, then  the output of SRNN layer summarizes the stimulus from all directions $\mathcal{D}$:
\begin{equation} \label{eq2}
  \mathbf{m}_{t}=\sum_{r\in\mathcal{D}}\mathbf{V}^r\mathbf{s}_{t}^r
\end{equation}
where $\mathbf{m}_{t}$  denotes the  output nodes of SRNN. When traversing the graph $\mathcal{G}_t$, the learned parameters $\{\mathbf{U}^r, \mathbf{W}^r, \mathbf{b}^r, \mathbf{V}^r\}$  are recurrently utilized  and they are not related with time. Such a process is designed as a network layer called SRNN layer to seamlessly connect other layers.

In SRNN layer, we consider four traversing directions starting from four angular points to make the traversing information from traversing processes mutually complementary. For example, the directional traversing from the top-left corner aims to capture contextual cues about the top-left areas with the adjacent predecessor set  $\mathcal{N}_{ij} =\{(i,i-1),(i-1,j-1),(i-1,j)\}$. Thus, four directed acyclic chains can be generated to represent the 2D neighborhood system by connecting contiguous elements and traversing these elements respectively from four directions. By doing this, discriminative spatial dependencies for emotion recognition can be modeled.

The representations learned from the SRNN layer are sequentially concatenated at each time slice and thus form a temporal sequence. For an entire emotion process, a single slice cannot reflect the characteristic of emotion due to its small granularity. The better strategy is to build the entire dynamic process rather than isolating considering each slice. RNNs can adaptively model such a temporal dynamic process. Here we employ a bi-directional RNN to simultaneously capture forward and backward dynamic transforms of sequence, i.e., two RNNs are respectively used to traverse the temporal sequence in a forward or backward behavior. Formally, suppose that sequential representations are denoted as $\mathbf{m}_t$ and the temporal length is $L$, then the TRNN layer can be written as
\begin{equation} \label{eq3}
  \mathbf{h}_{t}^{f}=\sigma_1(\mathbf{W}_{ih}^f\mathbf{m}_{t}+\mathbf{W}_{hh}^f\mathbf{h}^f_{t-1}+\mathbf{b}^f),
\end{equation}
\begin{equation} \label{eq4}
  \mathbf{h}_{t}^{b}=\sigma_1(\mathbf{W}_{ih}^b\mathbf{m}_{t}+\mathbf{W}_{hh}^b\mathbf{h}^b_{t-1}+\mathbf{b}^b),
\end{equation}
where $\{\mathbf{W}_{ih}^f, \mathbf{W}_{hh}^f,  \mathbf{b}^f\}$  and  $\{\mathbf{W}_{ih}^b, \mathbf{W}_{hh}^b,  \mathbf{b}^b\}$ are the learned parameters for recurrently traversing the sequences scanned forward and backward respectively, $\mathbf{m}_{t}, \mathbf{h}_{t}^f$ and $\mathbf{h}_{t}^b$ are the input nodes, hidden nodes for the forward scanned network  and hidden nodes  the backward scanned network respectively.  Similar to spatial projection in SRNN layer, projection matrices  denoted as  $\mathbf{G}^{f}=[G^f_{ij}]_{L \times L_p}$ and $\mathbf{G}^{b}=[G^b_{ij}]_{L \times L_p}$ are also applied to the temporal hidden states, resulting in the following expressions:
\begin{equation} \label{tmpro}
  \mathbf{q}^f_{t}=\sum_{i=1}^LG^f_{it}\mathbf{h}_{i}^f,~~\mathbf{q}^b_{t}=\sum_{i=1}^LG^b_{it}\mathbf{h}_{i}^b, ~~~t=1,\cdots,L_p
\end{equation}
where $L_p$ denotes the length after temporal projection and $\mathbf{q}^f_{t}$, $\mathbf{q}^b_{t}$ respectively denote the $t$th column vectors of the output matrices of the forward and backward scanned networks.

Let $$ \mathbf{q}^f=[(\mathbf{q}^f_{1})^T, \cdots,  (\mathbf{q}^f_{L_p})^T]^T$$ and $$\mathbf{q}^b=[(\mathbf{q}^b_{1})^T, \cdots,  (\mathbf{q}^b_{L_p})^T]^T$$  denote the concatenated vectors for the  forward and backward scanned networks respectively. Then, the output of  TRNN layer denoted as  $\mathbf{o}$ can be calculated by the following equation
\begin{equation} \label{eq5}
  \mathbf{o}=\mathbf{V}^f \mathbf{q}^f+\mathbf{V}^b \mathbf{q}^b,
\end{equation}
where $\mathbf{o}=[{o}_{1}, {o}_{2}, \cdots, {o}_{C}]^T$ and $C$ equals the number of emotion types.

Finally, the output nodes of TRNN layer are fed into the softmax layer for emotion classification:
\begin{equation} \label{eq5}
        P(i|\mathbf{X})=\exp({o}_{i})/\sum_{k=1}^C{\exp({o}_{k})},
\end{equation}
where $P(i|\mathbf{X})$ denotes the probability for the input $\mathbf{X}$ being predicted as the $i$th class.

In addition, we use cross entropy loss defined as follows to represent the objective loss function, which can be written as
\begin{eqnarray} \label{eq5}
\nonumber E =-\sum_{i=1}^N\sum_{c=1}^C\tau(y_i,c) \times \log P(c|\mathbf{I}_i)+ \lambda_1\sum_{r\in\mathcal{D}}\sum_{i=1}^{K_p}||\mathbf{g}^{r}_{i}||_1  \\
            +\lambda_2(\sum_{i=1}^{L_p}||\mathbf{g}^f_i||_1+\sum_{i=1}^{L_p}||\mathbf{g}^b_i||_1),
\end{eqnarray}
in which
\begin{equation} \label{eq4}
   \nonumber   \tau(y_i,c)=
      \begin{cases}
         1,  &  \mbox{if} ~y_i=c; \\
         0,  &  \mbox{otherwise.}
       \end{cases}
\end{equation}
where $E$ denotes the cross entropy loss, $N$ denotes the number of the training samples, $\mathbf{I}^i$ represents the $i$th training sample of the training set,  $y_{i}$ is the label of the $i$th training sample, $\mathbf{g}^{r}_{i}$, $\mathbf{g}^f_i$, $\mathbf{g}^b_i$ denote the $i$th column vectors of $\mathbf{G}^{r}$, $\mathbf{G}^f$, $\mathbf{G}^b$ respectively.

In the loss function, the first term calculates the mean negative logarithm value of the prediction probability of the training samples. The second and third terms ensure the sparse structure of the matrices in  spatial and temporal projection.  As elements of  projection matrices indicate the importance of the corresponding  spatial or temporal hidden states, the  sparse structure is able to endow high weights to the discriminative hidden states while low weights to others, which achieves the purpose of selecting salient hidden states.


The proposed STRNN can be effectively optimized by the classic back propagation through time (BPTT) algorithm. In BPTT, the recurrent nets can be converted into common feed-forward networks after they are unfolded to a sequence with a limited size. Thus, traditional gradient back-propagation used in common deep networks can be directly applied.

\section{Experiments}
\label{Exp}

In this section, we firstly introduce the datasets we use for testing the performance of our proposed STRNN, then report and analyze the results of our method on these datasets by comparing with other state-of-the-art methods.

\subsection{Datasets and feature extraction}

The proposed STRNN method is tested on both SJTU Emotion EEG Dataset (SEED)~\cite{zheng2015investigating} and the dataset of CK+ facial expression image sequences~\cite{lucey2010extended}. SEED  contains three categories of emotions (positive, neutral, and negative) of fifteen subjects (7 males and 8 females), which are elicited by showing emotional film clips to the participants. The EEG signals of these subjects were recorded  using an ESI NeuroScan System at a sampling rate of 1000 Hz from 62-channel electrode cap according to the international 10-20 system. The CK+ dataset consists of  327 image sequences with seven emotion labels: anger (An), contempt (Co), disgust (Di), fear (Fe), happiness (Ha), sadness (Sa), and surprise (Su) of 118 subjects. In this database, each sequence starts with a neutral emotion and ends with a peak of the emotion.

\begin{figure}[!h]
  \centering
  \includegraphics[height=2.4in, width=3.3in]{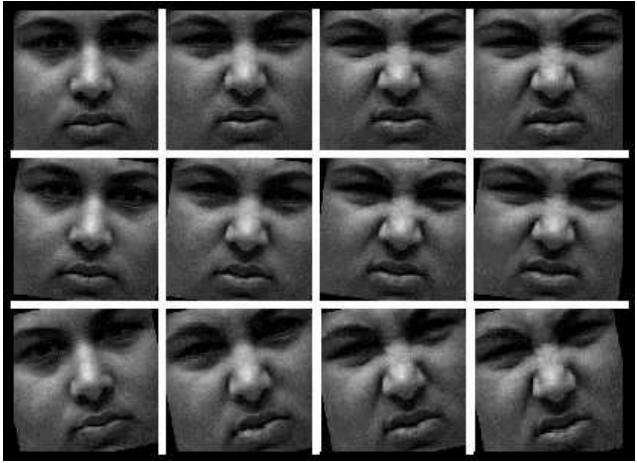}
  \caption{Samples of data augmentation of CK+. The first row contains four original frames sampled from a sequence, and the second and third rows contain the images which are rotated 7$^o$ clockwise and 12$^o$  counterclockwise corresponding to the images in the first row. }
  \label{CK+samp}
\end{figure}

\begin{figure*}[!t]
  \centering
  \includegraphics[height=2.2in, width=4.8in]{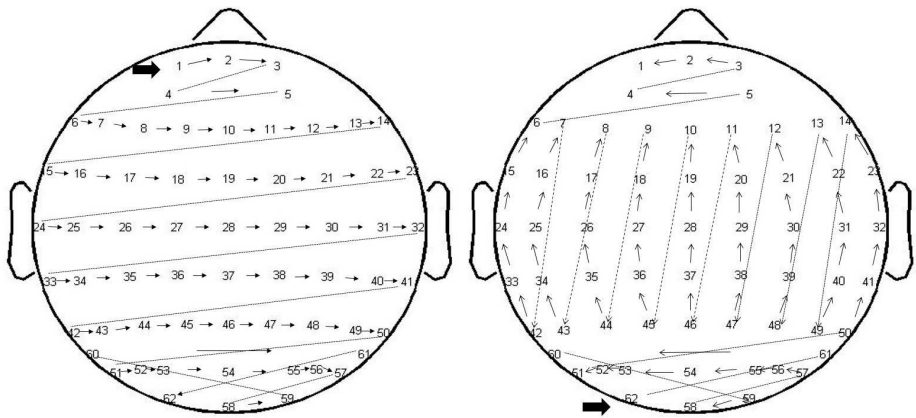}
  \caption{The scanning order of electrodes in two directions. The other two scanning directions inverse the current scanning orders. }
  \label{scanorder}
\end{figure*}

\begin{table*}[!t]
\renewcommand{\arraystretch}{1.3}
\caption{The comparisons on EEG signal based emotion dataset SEED.}
\label{SEEDresult}
\centering
\begin{tabular}{|c|c|c|c|c|}
\hline
Feature          & classifier                              & number of channels                       & \tabincell{c}{ number of  \\ frequency bands}          & accuracy ($\%$) \\
\hline
    DE           & SVM \cite{zheng2015investigating}       &      62                                  & 5                                                       &  83.99\\
\hline
    DE           & SVM \cite{zheng2015investigating}       &      12                                  & 5                                                       &  86.65\\
\hline
    DE           & LDA \cite{duda2012pattern}              &      62                                  & 5                                                       &  80.32 \\
\hline
    DE           & CCA  \cite{hardoon2004canonical}        &      62                                  & 5                                                       &  76.16\\
\hline
    DE           & GSCCA \cite{zhengmultichannel}          &      62                                  & 5                                                       &  82.35\\
\hline
    DE           & DBN \cite{zheng2015investigating}       &      62                                  & 5                                                       &  86.08\\
\hline
    DE           & STRNN                                   &      62                                  & 5                                                       & \textbf{89.50}\\
\hline
\end{tabular}
\end{table*}

To recognize emotion from EEG signals, differential entropy (DE) descriptors~\cite{zheng2015investigating} are extracted, which are calculated in five frequency bands (delta: 1-3 Hz, theta: 4-7 Hz, alpha: 8-13 Hz, beta: 14-30 Hz, gamma: 31-50 Hz) of 62 channels. For a specified continuous EEG sequence, a 256-point Short-Time Fourier Transform with a nonoverlapped Hanning window of 1s is used to extract five frequency bands of EEG signals and differential entropy is calculated for each frequency band. After this process, discrete sequences in five bands of 62 channels are generated. Then we use a slicing window of 9s  to temporally scan the  sequences by one step. For each step, the sequences in the slicing window are used as the  representation of the point which is in the center of the slicing window. By doing this, the temporal dependencies can be involved while recognizing the human emotion at a specific moment. This is quite different from \cite{zheng2015investigating} which just focuses on recognizing the average energy within a short time ignoring the temporal variation information.
For CK+, we use the pre-trained model proposed in \cite{parkhi2015deep} to extract features in each image to reduce the effects of noises or variant face poses so as to improve the representation ability. As the number of the training samples is limited, we perform rotation transformation to the sequences to achieve data augmentation. Each image is rotated with  angles including $7^\circ$,  $-7^\circ$, $12^\circ$ and  $-12^\circ$ so that there are totally 1635 samples. Some examples of this rotation process are shown in Fig. \ref{CK+samp}. For each image, the feature maps of the pooling layer, which is located before the first fully connected layer of the pre-trained model, are used as the representation.  For each input image sequence, the extracted feature maps are concatenated temporally as the representation of this sequence.

\subsection{EEG Signals Based Emotion Recognition }

The basic experiment configuration is the same to the one in~\cite{zheng2015investigating}. In this dataset, there are total fifteen subjects and each subject is conducted with the experiments across two time sessions. Thus there are totally 30 experiments evaluated here. Following the same protocol in~\cite{zheng2015investigating}, the training data and the testing data are respectively taken from different sessions of the same experiment. There are nine sessions for training and the remaining six sessions for testing.

In SRNN layer, the numbers of the input, hidden and output nodes are set to be 5, 30 and  30 respectively, and the number of hidden states ($K_p$ in Eqn. (\ref{sppro})) is reduced from 62 to 10  after spatial projection. In TRNN layer, the numbers of hidden and output nodes are set to be 30 and 3, where the number of the output nodes is set according to the number of emotion types. The number of hidden states ($L_p$ in Eqn. (\ref{tmpro})) is reduced from 9 to 5 after temporal projection. These parameters of our STRNN are roughly set without {\color{blue} elaborate traversal}.  In SRNN layer, the RNNs scan the electrodes from four angels. As the distribution of locations of the electrodes is not exactly a rectangle, we define the scanning order as shown in Fig. \ref{scanorder} to model intimate interactions existing among those spatially adjacent electrodes.

\begin{table*}[!t]
\renewcommand{\arraystretch}{1.3}
\caption{The performance of different frequency bands on SEED.}
\label{SEEDfreqresult}
\centering
\begin{tabular}{|c|c|c|c|c|c|c|}
\hline
frequency band                      & Delta                      & Theta                    & Alpha                          & Beta                     &  Gamma                  & all \\
\hline
DBN \cite{zheng2015investigating}   & 64.32/12.45                &  60.77/10.42             &  64.01/15.97                   &  78.92/12.48             &  \textbf{79.19/14.58}   &  86.08/8.34 \\
\hline
STRNN                               &   \textbf{80.9/12.27}      &  \textbf{83.35/9.15}     &  \textbf{82.69/12.99}          &  \textbf{83.41/10.16}    &  69.61/15.65            &  \textbf{89.50/7.63} \\
\hline
\end{tabular}
\end{table*}

\begin{figure}[!h]
  \centering
  \includegraphics[height=1.3in, width=2.0in]{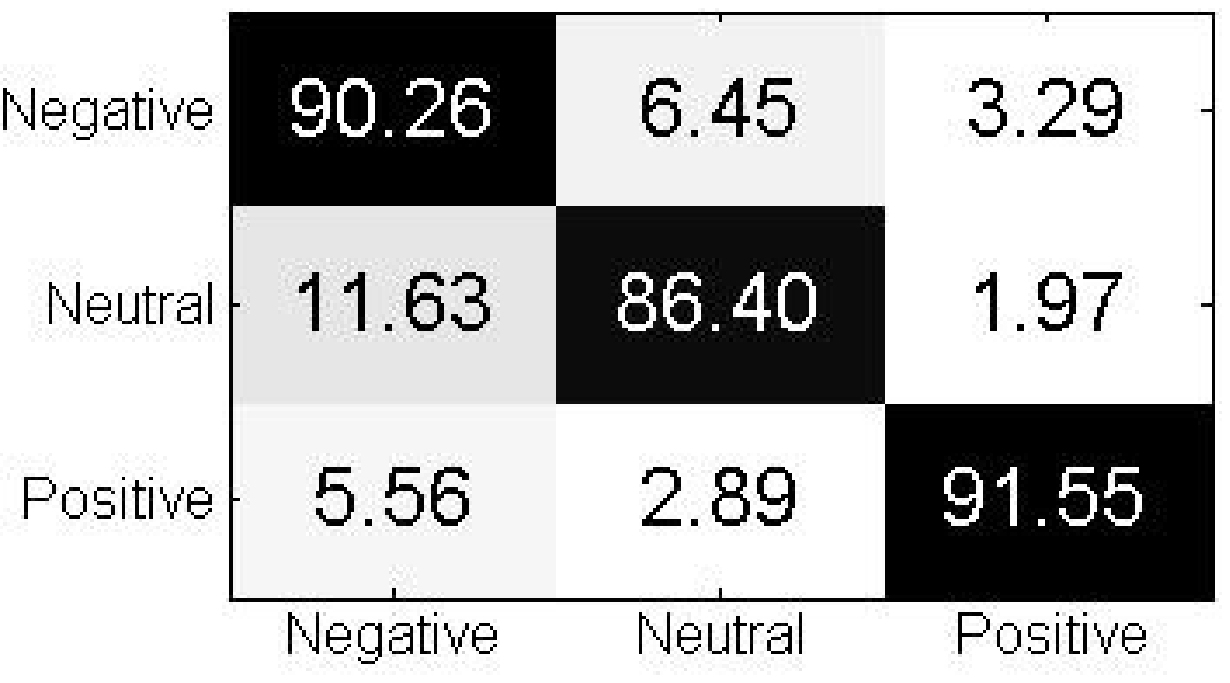}
  \caption{The experimental results of confusion matrix on SEED. }
  \label{EEGconf}
\end{figure}
 The average accuracy of STRNN with the DE features of all frequency bands in 30 experiments of fifteen subjects is shown in Table \ref{SEEDresult}.  This result is also compared with various existed algorithms under the same protocol of nine sessions for training and the remaining six sessions for testing, where these algorithms include linear discriminant analysis (LDA) \cite{duda2012pattern}, canonical correlation analysis (CCA) \cite{hardoon2004canonical}, SVM, DBN and so on. Most of these methods employ  DE features of all 62 channels except SVM \cite{zheng2015investigating}, which uses both 62 and 12 channels. SVM of 62 channels achieves the accuracy of 83.99$\%$ while LDA achieves the accuracy of 80.32$\%$.  The performance of SVM is further improved to be 86.65$\%$ by selecting certain 12 channels out of full 62 channels. CCA gets the performance of 76.16$\%$ while GSCCA \cite{zhengmultichannel} achieves much higher accuracy of 82.35$\%$ by endowing conventional CCA with the ability of handling the group feature selection problem from raw EEG features. DBN achieves the accuracy of 86.08$\%$ which is highest among the existed methods using DE features of full 62 channels. Our STRNN achieves the accuracy of 89.50$\%$ which is 3.42 percent higher than DBN. This performance gain indicates that our STRNN benefits from modeling the spatial and temporal dependencies layer by layer while DBN just concatenates them together without considering the spatio-temporal structure of the EEG signals.

To reveal which frequency oscillation of brain activity is more related to the emotion processing, the performance of the DE feature on different frequency bands (Delta, Theta, Alpha, Beta and Gamma) are compared between DBN and STRNN, which is shown in Table \ref{SEEDfreqresult}. As we can see, the distribution of the accuracies of STRNN on different frequency bands is quite different from the result of DBN: the accuracies achieved on  four frequency bands ( Delta, Theta, Alpha and Beta ) are all more than 80$\%$ while the accuracy of Gamma is lower. The highest accuracy  is achieved on  Beta band. However, for the results of DBN in  \cite{zheng2015investigating}, only beta and gamma bands of EEG signals  are more related with emotion processing than other frequency bands. This difference may be caused by the fact that the temporal slicing window we use during the feature extraction process involves temporal dependencies. And according to our results, the spatial-temporal dependencies of the four frequency bands, i.e. Delta, Theta, Alpha and Beta, contribute more to the recognition of emotion. Moreover, the deviations of recognition results of all five bands are calculated as well as those of each specific frequency band. The values of  deviations of STRNN are lower than DBN except Gamma band, which indicates the performance of our STRNN is more stable across different experiments of different subjects.

Fig. \ref{EEGconf}  shows the confusion matrix of all evaluated experiments for SEED. As it is shown,  our algorithm performs well in recognizing all three types of emotion as the accuracies of them are more than  85.0$\%$.  Positive and negative emotion are easier to be recognized  whereas neutral, by contrast,  is relatively difficult to be correctly classified as it is easily confused with negative.

\subsection{Video Emotion Recognition}
In this experiment, we train and test the CK+ database with the 10-fold cross validation by following the previous protocol. The database is divided into 10 subsets by  ID of the subjects in ascending order, where the subjects in any two of subsets are not overlapped. For each run, 9 subsets were employed for training and the remaining one subset for testing. Such 10 runs are performed by enumerating the subset used for testing  and the average recognition performance is computed as the final result of the 10 runs.

  The parameters of our STRNN are set as follows: the numbers of the input, hidden and output nodes in  SRNN layer are set to be 512, 50 and  50 respectively, the number of hidden states ($K_p$ in Eqn. (\ref{sppro})) is reduced from 49 to 10 after spatial projection. In  TRNN layer, the numbers of hidden and output nodes are set to be 150 and 7.  The number of hidden states ($L_p$ in Eqn. (\ref{tmpro})) is reduced from 44 to 5  after temporal projection.

Many state-of-the-art methods adopt different protocols  for CK+ are compared with our algorithm, including  Cov3D \cite{sanin2013spatio}, TMS \cite{jain2011facial}, STM-ExpLet \cite{liu2014learning} and so on which are shown in Table \ref{CK+acc}. Most of these methods adopt  10-fold cross validation except Cov3D and TMS, which adopt 5-fold and 4-fold cross validation respectively. 3D SIFT based method achieves the accuracy of 81.4$\%$ while 3D HOG and MSR achive the accuracy of 91.4$\%$. 3D CNN only gets the accuracy of 85.9$\%$ while 3DCNN-DAP can achieve 92.4$\%$ which benefits from using strong spatial structural constraints on the dynamic action parts. Deep temporal appearance network (DTAN)  employed in \cite{jung2015joint} achieves the accuracy of 91.4$\%$ by applying a CNN model which is able to capture temporal changes of appearance. The STM-ExpLet achieves the best performance of 94.2$\%$ among the compared methods by introducing complex manifold structures. Our STRNN achieves 95.4$\%$ which is more competitive to these state-of-the-art methods.

\begin{figure}[!h]
  \centering
  \includegraphics[height=2.2in, width=2.8in]{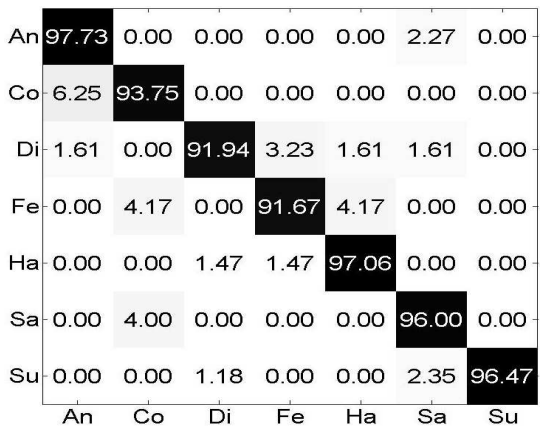}
  \caption{The experimental results of confusion matrix on  CK+.}
  \label{CK+conf}
\end{figure}

Fig.~\ref{CK+conf} shows the confusion matrix for CK+. In general, our algorithm performs well in recognizing all types of emotion as the accuracy of each expression is more than  90$\%$. Among them, four kinds of expressions including  anger, happiness, sadness and surprise are relatively easy to be recognized with the accuracies of 97.73$\%$, 97.06$\%$, 96.00$\%$ and 96.47$\%$, which may be attribute to their relatively large muscle deformations. Next ones are contempt, disgust and  fear, respectively with 93.75$\%$, 91.94$\%$ and 91.67$\%$ recognition rates.  Relatively high confusions appear between three pairs of expressions, contempt versus angry, contempt versus fear and contempt versus sadness, which may be intuitively due to the similar  muscle deformations.

\begin{table}[!h]
\renewcommand{\arraystretch}{1.3}
\caption{The comparisons on video face based emotion dataset CK+.}
\label{CK+acc}
\centering
\begin{tabular}{|c|c|c|}
\hline
 method                                          &  cross validation protocol                        & accuracy($\%$) \\
\hline
      3D SIFT \cite{scovanner20073}              &  10-fold                                          & 81.4\\
\hline
      3D HOG \cite{klaser2008spatio}             &  10-fold                                          & 91.4\\
\hline
      MSR \cite{ptucha2011manifold}              &  10-fold                                          & 91.4 \\
\hline
      Cov3D \cite{sanin2013spatio}               &  5-fold                                           & 92.3\\
\hline
      STM-ExpLet \cite{liu2014learning}          &    10-fold                                        & 94.2\\
\hline
      TMS \cite{jain2011facial}                  &    4-fold                                         & 91.9\\
\hline
      3DCNN \cite{liu2014deeply}                 &  10-fold                                          & 85.9\\
\hline
     3DCNN-DAP \cite{liu2014deeply}              &   10-fold                                         &92.4\\
\hline
     DTAN \cite{jung2015joint}                  &   10-fold                                         &91.4\\
\hline
     STRNN                                       &   10-fold                                         &\textbf{95.4}\\
\hline
\end{tabular}
\end{table}

\section{Conclusions and Discussions}
\label{conclu}
In this paper, a novel STRNN method is proposed to deal with EEG signal based and face image based human emotion recognition. To well model spatial co-occurrent variations and time dependent variation of human emotions, a multi-direction SRNN layer and a bi-direction TRNN layer are hierarchically employed to learn spatial and temporal dependencies layer by layer. To adapt the multi-channel EEG signals to the proposed STRNN framework, the spatial scanning order of electrodes are specified by spatial coordinates and temporal variation information is involved by slicing a window on the extracted DE feature sequences.   To further select those salient regions of emotion representation as well as increase the model discriminant ability, we impose sparse projection onto those hidden states of spatial and temporal domains.  The experimental results on both SEED EEG dataset and CK+ facial expression dataset have demonstrated that the proposed STRNN method achieves the state-of-the-art performance. 

As the STRNN method can be seen as an integration of both SRNN and TRNN, it is still interesting to see how much improvement could be obtained by RNN modeling in the spatial or temporal domains or how much performance improvement can be gained by adding sparse constraints in STRNN. In addition, it is also interesting to see what the salient regions learned by the sparse STRNN would be located in the facial expression images. To answer all of these questions, we will also conduct additional experiments in what follows.

\subsubsection{Comparisons of STRNN with SRNN and TRNN}
To compare STRNN with SRNN and TRNN, the STRNN method is revised into only SRNN and only TRNN, in which process the other spatial or temporal RNN is merged into a full connection layer. The results are shown in Table \ref{STvsSvsT}.
\begin{table}[!h]
\renewcommand{\arraystretch}{1.3}
\caption{The results of SRNN, TRNN and STRNN on SEED .}
\label{STvsSvsT}
\centering
\begin{tabular}{|c|c|c|c|}
\hline
 Model                  &\tabincell{c}{ number of \\ channels}     & \tabincell{c}{ number of  \\ frequency bands}   & accuracy ($\%$) \\
\hline
SRNN                    &      62                                  & 5                                               &  85.88/9.98\\
\hline
TRNN                    &      62                                  & 5                                               &  85.20/9.13\\
\hline
 STRNN                  &      62                                  & 5                                              & \textbf{89.50/7.63}\\
\hline
\end{tabular}
\end{table}
The network which contains only TRNN achieves the accuracy of 85.20$\%$ with the deviation of 9.13$\%$, while the network containing only SRNN achieves a little higher accuracy of 85.88$\%$ but with a higher deviation of 9.98$\%$. STRNN achieves the accuracy of 89.50$\%$ which is about 4$\%$ higher than SRNN or TRNN  with a lower deviation. The improvement of performance demonstrates the effectiveness of the hierarchical structure of spatial and temporal RNN layers which learns both spatial and temporal dependencies.

\subsubsection{Comparisons of STRNN with non-sparse STRNN}
To Compare STRNN with non-sparse STRNN, we conduct experiments using the proposed  STRNN method and an  STRNN method without sparse constraints on projection matrices. The results are shown in Table \ref{SPvsnSP}.
\begin{table}[!h]
\renewcommand{\arraystretch}{1.3}
\caption{The results of STRNN and non-sparse STRNN.}
\label{SPvsnSP}
\centering
\begin{tabular}{|c|c|c|}
\hline
 Dataset                    & ~~~~Method~~~~ & ~~Accuracy~~ \\
\hline
\multirow{2}{*}{SEED}       &     non-sparse STRNN       & 88.1 \\
\cline{2-3}      {}         &      STRNN       & \textbf{89.5} \\
 \hline
\multirow{2}{*}{CK+}     &     non-sparse  STRNN       & 94.2 \\
\cline{2-3}      {}          &      STRNN       & \textbf{95.4} \\
\hline
\end{tabular}
\end{table}
As we can see, for both SEED and CK+ datasets, the accuracies of STRNN are about more than one percent higher than those of non-sparse STRNN, which verifies the effectiveness of sparse constraints which improve the performance of the proposed  STRNN as well as achieve salient emotion regions detection.

\subsubsection{Salient emotion detection}

\begin{figure}[!h]
  \centering
  \includegraphics[height=1.3in, width=2.5in]{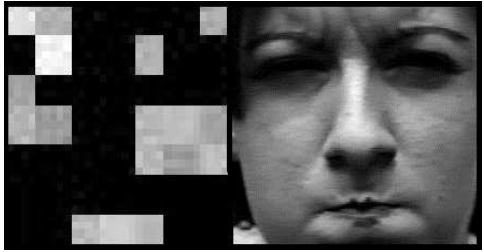}
  \caption{ Example of the weight distribution over facial regions. }
  \label{fawei}
\end{figure}

In addition to showing  average recognition accuracies, we also visualize the weights of hidden states  of the multi-direction SRNN layer in our STRNN in the experiment conducted on  CK+ dataset.  In this process, the columns of the absolute coefficient values of projection matrices $\mathbf{G}^r$ are averaged over all spatial traversing directions.  Fig. \ref{fawei} shows the distribution of detected salient facial regions by mapping the weights of hidden states back to corresponding spatial regions in a 2D facial image.
 As it is shown, the highlighted parts in the left image of  Fig. \ref{fawei} correspond to the action regions around mouth, eyes and nose, which are intuitively crucial for human to perceive facial expression. Moreover, the black regions in Fig. \ref{fawei} indicate that most values in projection matrices are near zero, which verifies the effectiveness of  $l_1$-norm terms in the loss function for ensuring the sparsity of the column vectors of projection matrices.



\ifCLASSOPTIONcaptionsoff
  \newpage
\fi



\end{document}